\documentclass{article}

\usepackage{microtype}
\usepackage{graphicx}
\usepackage{subcaption}
\usepackage{booktabs}

\usepackage{hyperref}

\usepackage[preprint]{icml2026}

\usepackage{amsmath}
\usepackage{amssymb}
\usepackage{mathtools}
\usepackage{amsthm}
\usepackage{listings}
\usepackage{xcolor}
\usepackage{tikz}
\usetikzlibrary{shapes,arrows.meta,positioning,fit}
\usepackage{enumitem}
\usepackage{tabularx}

\usepackage[capitalize,noabbrev]{cleveref}

\theoremstyle{plain}
\newtheorem{theorem}{Theorem}[section]

\theoremstyle{definition}
\newtheorem{definition}[theorem]{Definition}

\theoremstyle{remark}

\makeatletter
\newenvironment{acks}
{
    \section*{Acknowledgments}
    \@mkboth{Acknowledgments}{Acknowledgments}
}
{}
\makeatother

\icmltitlerunning{QED: An Open-Source Multi-Agent System for Generating Mathematical Proofs on Open Problems}

\begin{document}

\twocolumn[
    \icmltitle{QED: An Open-Source Multi-Agent System for Generating \\ Mathematical Proofs on Open Problems}


    \icmlsetsymbol{equal}{*}

    \begin{icmlauthorlist}
        \icmlauthor{Chenyang An}{ucsd}
        \icmlauthor{Qihao Ye}{ucsd}
        \icmlauthor{Minghao Pan}{caltech}
        \icmlauthor{Jiayun Zhang}{aws}
    \end{icmlauthorlist}

    \icmlaffiliation{ucsd}{University of California, San Diego, La Jolla, USA}
    \icmlaffiliation{caltech}{California Institute of Technology, Pasadena, USA}
    \icmlaffiliation{aws}{Amazon Web Services (work done outside of Amazon)}

    \icmlcorrespondingauthor{Chenyang An}{c5an@ucsd.edu}

    \icmlkeywords{Machine Learning, Mathematical Reasoning, Multi-Agent Systems}

    \vskip 0.3in
]

\printAffiliationsAndNotice{}

\begin{abstract}
    We present \textbf{QED}, an open-source multi-agent system that turns human-provided research questions into complete mathematical proofs without further human guidance. Its pipeline is designed to overcome common failures of single-query proof generation by separating planning, proving, and verification: a decomposition agent structures the proof search, prover agents generate candidate arguments, and verifier agents check correctness.
    In collaboration with domain experts, we evaluated QED on 18 research-level projects of varying difficulty. QED produced five original works across algebraic geometry, fluid PDEs, probability, and inverse problems. Expert assessments regard these works as solid specialized research contributions, with three comparable in difficulty and scope to work commonly published in established specialist mathematics venues. QED is released at \url{https://github.com/proofQED/QED}.
\end{abstract}

\section{Introduction}
Large language models (LLMs) have demonstrated strong and rapidly improving performance on mathematical reasoning, progressing from grade-school word problems~\cite{cobbe2021trainingverifierssolvemath} and competition mathematics~\cite{hendrycks2021measuringmathematicalproblemsolving,trinh2024solving,Guo_2025} toward research-level use. Recent reports include individual mathematicians using LLM chatbots to obtain core ideas for ongoing research~\cite{gowers2026chatgpt,alexeev2026primitivesets}, as well as large-scale systems such as Aletheia~\cite{feng2026autonomousmathematicsresearch}, which have contributed to mathematical discovery with varying degrees of human guidance. Fully automatic proof generation from only a human-provided problem statement remains rarer. Notable examples include Aletheia’s autonomous results, including~\cite{feng2026eigenweights}, OpenAI's short solutions to Erdős problems~\cite{alexeev2026shortproofs,alexeev2026shortproofs2} and OpenAI’s recent announcement that an internal reasoning model produced a counterexample to a longstanding conjecture in the planar unit-distance problem~\cite{openai2026planarunitdistanceproof}. Google DeepMind's AI Co-Mathematician further explores human-in-the-loop mathematical research through a Gemini-based, stateful multi-agent workbench~\cite{zheng2026ai}. These developments mark a shift from benchmark problem solving toward AI-assisted mathematical research. However, most high-profile demonstrations still rely on closed systems, models that are not publicly released, or human-directed workflows. This motivates \textbf{QED}, an open-source, fully automatic multi-agent proof system evaluated on expert-proposed research problems.

Through experiments with frontier coding agents (Claude Code, Codex, and Gemini CLI) on research-level proof tasks, we observe that naive deployment (single-turn prompting, unchecked outputs, no structured verification) leads to identifiable failure modes (Section~\ref{sec:failure_modes}).
We characterize seven such failure modes and demonstrate that each can be mitigated through a targeted architectural intervention.
Each architectural design choice in QED is directly motivated by an observed failure mode:
\begin{enumerate}[leftmargin=*,itemsep=2pt]
    \item \textbf{Context contamination}: Prover and verifier share context, producing circular reasoning $\rightarrow$ \emph{separated prover and verifier agents}.

    \item \textbf{Citation hallucination}: Models fabricate or misstate referenced theorems $\rightarrow$ \emph{structured citation verification with exact-statement matching}.

    \item \textbf{Misallocation of proof effort}: Models skip the hardest parts of proofs while continuing to work on easy steps $\rightarrow$ \emph{mandatory \texttt{<key-original-step>} tagging with verifier enforcement}.

    \item \textbf{Unstable proof plans}: Models abandon strategies mid-proof on hard problems $\rightarrow$ \emph{decomposition mode separating proof planning from proof execution}.

    \item \textbf{Unfocused verification}: Models dilute attention across heterogeneous checks $\rightarrow$ \emph{two-stage verification (structural then detailed)}.

    \item \textbf{Problem modification}: Models alter the problem statement to reduce difficulty $\rightarrow$ \emph{problem-statement integrity checking (word-by-word comparison)}.

    \item \textbf{Single-model bottleneck}: Relying on one model makes the system vulnerable to that model's blind spots $\rightarrow$ \emph{multi-model parallel proving and brainstorm mode}.
\end{enumerate}

Through our collaboration with seven domain experts, we asked experts to propose research projects for which, to the best of their knowledge, no proof or complete solution was known, and whose successful resolution would constitute a publishable contribution.
We collected 18 research projects and tested QED on them.
QED successfully produced original, nontrivial, and complete solutions to 5 research projects in algebraic geometry, fluid PDE, probability, and inverse problems.
QED was operated fully automatically, and no expert guidance was provided during the runs.
Each proof was independently verified by the contributing domain experts, who confirmed the results to be \emph{original} (not previously known) and \emph{nontrivial} (requiring genuine mathematical insight).
Moreover, the observed false positive rate of the QED verifier with Codex GPT-5.5 in our experiment was 0 out of 17.
In particular, across 214 proof candidates produced by the QED prover using Codex GPT-5.5 and evaluated by the QED verifier under the same configuration, 17 candidates were accepted by the verifier.
Every verifier-accepted candidate was subsequently accepted by the corresponding domain expert, yielding an observed false-positive rate of 0 in this setting.

Links to the problem statements, proofs, and expert commentary for the five research projects that QED successfully solved can be found in the appendix~\ref{app:pos}.

Note that we do not report full details of all the unsuccessful cases for QED because some of the research problems remain active, with human mathematicians still working on them. Many of the unsuccessful cases are longstanding open problems that are widely regarded as genuinely difficult, and we share some of them in the appendix~\ref{app:neg}.

\noindent In summary, our contributions are as follows:
\begin{itemize}[leftmargin=*,itemsep=2pt]
    \item We give an empirical characterization of seven failure modes of frontier LLMs on research-level proof tasks (Section~\ref{sec:failure_modes}).

    \item We design and release \textbf{QED}, an open-source multi-agent proof system that fully automates the theorem-proving process given human-provided problem statements, with each architectural decision driven by a specific failure mode (Section~\ref{sec:system}).

    \item Through collaborations with domain experts, we show that QED independently produced original, nontrivial proofs for five open research projects across probability, algebraic geometry, fluid PDEs, and inverse problems, with all resulting proofs verified and validated by the relevant experts. Expert assessments further indicate that three of these results are comparable in difficulty and scope to work published in established specialist mathematics journals. We also report selected unsuccessful cases, providing direct evidence of both the capabilities and current limitations of LLMs when embedded in a well-designed multi-agent pipeline for open mathematical research problems.
\end{itemize}

\section{Related Work}
\label{sec:related_work}

\paragraph{LLM mathematical reasoning.}
The rapid improvement of LLMs on open-source mathematical benchmarks, from GSM8K~\cite{cobbe2021trainingverifierssolvemath} and MATH~\cite{hendrycks2021measuringmathematicalproblemsolving} to competition-level problems~\cite{trinh2024solving}, has made those benchmarks unable to measure SOTA models' reasoning capabilities. Chain-of-thought prompting~\cite{wei2023chainofthoughtpromptingelicitsreasoning}, self-consistency~\cite{wang2023selfconsistencyimproveschainthought}, training models on better data~\cite{an2024learnfailurefinetuningllms, an2025nexttokenpredictiontaskassumes}, and reinforcement learning from human feedback or reward models~\cite{Guo_2025,an2025derl} have all contributed to this progress. However, benchmark performance does not directly imply the ability to produce original research-level proofs, as benchmarks consist of problems with known solutions.
To address the saturation of open benchmarks, more recent efforts have introduced harder, often closed-source evaluations such as Frontier Math and the Humanity’s Last Exam benchmark, which aim to probe the limits of current models on substantially more challenging and less memorized problems~\cite{phan2025humanity, glazer2025frontiermathbenchmarkevaluatingadvanced}. However, these benchmarks are not publicly available, limiting reproducibility and systematic analysis.

\paragraph{Agent architectures for coding and scientific discovery.}
Recent multi-agent LLM systems increasingly move beyond simple chat-based collaboration toward structured agent architectures for complex technical and scientific workflows. In software engineering, systems such as MAAD~\cite{li2025maadautomatesoftwarearchitecture} and ALMAS~\cite{tawosi2025almasautonomousllmbasedmultiagent}  decompose development into specialized roles and stages, coordinating agents for architecture design, implementation, validation, and integration. In scientific discovery, systems such as MatSciAgent~\cite{chaudhari2026modular}, Robin~\cite{ghareeb2025robinmultiagentautomatingscientific}, DORA AI Scientist~\cite{naumov2025dora}, and MACC~\cite{oyama2026maccmultiagentcollaborativecompetition} organize agents around domain-specific scientific workflows, including hypothesis generation, literature analysis, experiment design, tool use, and report generation. More broadly, CORAL~\cite{qu2026coralautonomousmultiagentevolution} studies autonomous multi-agent evolution for open-ended discovery, emphasizing sustained search and knowledge accumulation across interacting agents~\cite{qu2026coralautonomousmultiagentevolution}.
QED differs from these systems in its specific focus on research-level mathematical proof, its failure-mode-driven design, and its multi-stage verification pipeline tailored to mathematical reasoning. In particular, QED explicitly targets challenges that are largely absent or less central in prior agent systems: rigorous citation grounding, preservation of problem-statement integrity, and formal consistency across a subgoal tree. Rather than relying only on role specialization, conversational refinement, or experimental feedback, QED introduces structured verification stages that enforce correctness at each level of the proof hierarchy, aligning the agent architecture with the logical and compositional nature of mathematical proof.

\paragraph{AI and open problems.}
Recent work has explored whether AI systems can contribute to open mathematical questions. FunSearch~\cite{romera2024mathematical} used LLMs to discover new constructions in extremal combinatorics. Aletheia~\cite{feng2026autonomousmathematicsresearch} demonstrated a closed-source agent that made research progress on open problems across multiple mathematical domains. OpenAI has also reported progress on Erd\H{o}s-style open problems: two manuscripts present eight short proofs across combinatorics, number theory, and probability, with the authors attributing the proofs to internal OpenAI models and describing the human role as digesting and polishing the writeups~\cite{alexeev2026shortproofs,alexeev2026shortproofs2}. Its recent planar unit-distance result similarly reports a counterexample to a longstanding conjecture produced by an internal reasoning model~\cite{openai2026planarunitdistanceproof}. Google DeepMind's AI Co-Mathematician is a limited-release Gemini-based workbench in which a project coordinator agent manages a stateful mathematical workspace and delegates tasks across parallel workstreams and specialized agents; the system is designed to support open-ended research workflows such as ideation, literature search, computational exploration, theorem proving, and theory building, and reports early tests in which it helped researchers solve open problems, identify new directions, and uncover overlooked references~\cite{zheng2026ai}. However, these results are either produced by closed-source agent or have faced scrutiny: some attributed solutions to open problems (e.g., those associated with Erd\H{o}s) were argued to be rediscoveries, and construction-focused results do not constitute proofs in the traditional sense. QED contributes to narrowing this gap as an open-source system producing natural-language proofs for genuinely open research problems, verified as original by domain experts.

\section{Why LLMs Fail at Real Proofs}
\label{sec:failure_modes}

Before introducing the architecture of QED, we characterize the failure modes that motivated its design. These observations emerge from experimentation with frontier coding agents (Claude Code, Codex and Gemini CLI) on research-level mathematical problems, prior to and during the development of QED.

\subsection{Failure Mode 1: Context Contamination}
\label{sec:fm1}
When a single LLM session serves as both prover and verifier, the verification step is contaminated by the prover's reasoning context. The model ``remembers'' its own proof strategy and is predisposed to accept its own arguments, even when they contain errors. This produces a false sense of correctness: the model generates a flawed proof, then verifies it as correct because both steps share the same internal state.

In our earliest attempts, we used the same LLM model to both generate and verify proofs, although in separate calls. This avoided literal context leakage, but still suffered from same-model self-verification bias: the verifier shared the prover’s reasoning tendencies and often failed to identify subtle flaws in proofs generated by that same model. This led to more frequent false positives and motivated our later design with a more independent verification stage.

\subsection{Failure Mode 2: Citation Hallucination}
\label{sec:fm2}
LLMs routinely fabricate mathematical citations. They invent theorem numbers, attribute results to wrong sources, fabricate URLs, or cite statements that do not appear in the referenced source. This is particularly dangerous in proof generation because a proof that relies on a hallucinated lemma may appear logically sound while being built on a nonexistent foundation.

In our experiments, citation hallucination was not a rare edge case. When no dedicated citation verifier was used, we observed that a high percentage of the citations produced by strong models such as Claude 4.6 Opus and GPT-5.5 in research-level proofs contained some form of hallucination or grounding error. These errors included fabricated theorem statements, incorrect attributions, nonexistent URLs, and citations to sources that did not actually contain the claimed statement. This made citation verification essential: without it, a proof could appear mathematically coherent while relying on unsupported or nonexistent lemmas.

\subsection{Failure Mode 3: Misallocation of Proof Effort}
\label{sec:fm3}
In our experiments, we found that LLMs tend to evade the core mathematical difficulties: while they often recognize what needs to be proven, they rely on vague language to bypass the rigorous execution required for genuine insight, resorting to phrases like “clearly, X holds” or “it is straightforward to verify” when the claim is neither clear nor straightforward. Meanwhile, they overinvest attention in elementary components, providing excessive detail to reprove well-established lemmas or justify routine steps; by exhausting their focus on these trivialities, they leave the primary, novel complexities of the problem largely underexplored.

\subsection{Failure Mode 4: Unstable Proof Plans}
\label{sec:fm4}
In our experiments, we found that LLMs typically form a proof plan before executing it, but they give up on plans too easily. When a model encounters a difficult step during execution, it tends to abandon the current plan and switch to an entirely different strategy rather than pushing through the difficulty. Over multiple iterations, this produces a random walk through proof space rather than systematic progress: the model cycles through superficially different plans without ever committing to one long enough to resolve its hard steps. The model fails to distinguish between ``this approach has a fixable error in execution'' and ``this approach is fundamentally flawed and needs a new plan.''

\subsection{Failure Mode 5: Unfocused Verification}
\label{sec:fm5}
When a single verification agent is tasked with checking everything---problem-statement integrity, citations, logical structure, and step-by-step correctness---it tends to dilute its attention across all aspects and miss critical issues. This has been identified repeatedly in our experiments.

\subsection{Failure Mode 6: Problem Modification}
\label{sec:fm6}
We also observed a form of reward hacking in proof generation: models sometimes subtly changed the problem being solved while still producing a plausible proof. For example, the model might change a quantifier, strengthen an assumption, weaken the conclusion, restrict the domain, or prove only a special case while presenting it as the original theorem. This is dangerous because the resulting proof may be mathematically correct, but for the wrong problem. In this sense, the model optimizes for the verifier’s proxy objective—producing something that looks like a valid proof—rather than satisfying the true objective of proving the exact user-specified statement. This is closely related to reward hacking or specification gaming, where a model exploits imperfections in the evaluation criterion to achieve a high score without accomplishing the intended task~\cite{bondarenko2025demonstratingspecificationgamingreasoning,wang2026rewardhackingeralarge}.

\subsection{Failure Mode 7: Single-Model Bottleneck}
\label{sec:fm7}
Different LLMs have different mathematical strengths and blind spots. A problem that one model cannot solve may be accessible to another. Relying on a single model means the system inherits all of that model's weaknesses, with no mechanism for complementary coverage.

\section{QED System Design}
\label{sec:system}

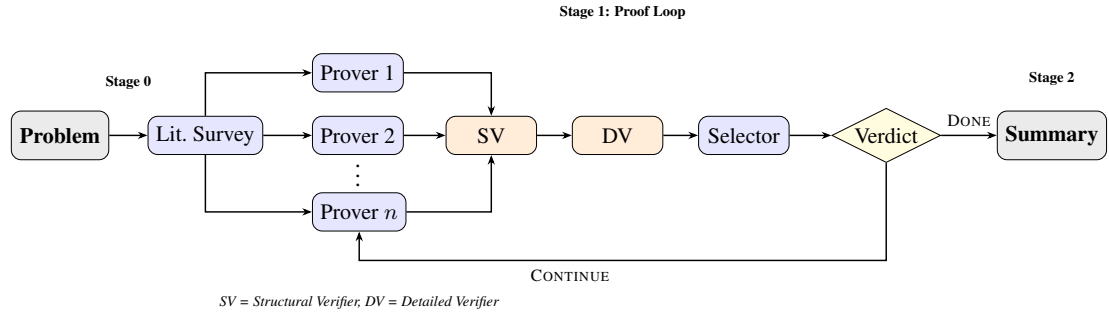
\begin{figure*}[t]
    \centering
    \resizebox{0.85\textwidth}{!}{%
        \begin{tikzpicture}[
                node distance=0.4cm and 0.55cm,
                stage/.style={draw, rounded corners, fill=black!8, minimum width=1.4cm, minimum height=0.7cm, align=center, font=\footnotesize\bfseries},
                agent/.style={draw, rounded corners, fill=blue!10, minimum width=1.3cm, minimum height=0.55cm, align=center, font=\footnotesize},
                verifier/.style={draw, rounded corners, fill=orange!15, minimum width=1.3cm, minimum height=0.55cm, align=center, font=\footnotesize},
                decision/.style={draw, diamond, fill=yellow!15, aspect=2, align=center, font=\footnotesize, inner sep=1pt},
                arr/.style={-{Stealth[length=4pt]}, semithick},
                darr/.style={-{Stealth[length=4pt]}, semithick},
                lbl/.style={font=\scriptsize, midway},
                fm/.style={font=\tiny\itshape, text=red!70!black},
            ]
            \node[stage] (input) {Problem};
            \node[agent, right=of input] (lit) {Lit.\ Survey};
            \draw[arr] (input) -- (lit);

            \node[agent, right=0.7cm of lit, yshift=0.9cm] (p1) {Prover 1};
            \node[agent, right=0.7cm of lit] (p2) {Prover 2};
            \node[below=-0.2cm of p2.south] (p23) {$\vdots$};
            \node[agent, right=0.7cm of lit, yshift=-1.1cm] (p3) {Prover $n$};
            \draw[arr] (lit) |- (p1);
            \draw[arr] (lit) -- (p2);
            \draw[arr] (lit) |- (p3);

            \node[verifier, right=0.6cm of p2] (sv) {SV};
            \draw[arr] (p1) -| (sv);
            \draw[arr] (p2) -- (sv);
            \draw[arr] (p3) -| (sv);

            \node[verifier, right=0.5cm of sv] (dv) {DV};
            \draw[arr] (sv) -- (dv);

            \node[agent, right=0.5cm of dv] (sel) {Selector};
            \draw[arr] (dv) -- (sel);

            \node[decision, right=0.6cm of sel] (v) {Verdict};
            \draw[arr] (sel) -- (v);

            \node[stage, right=0.8cm of v] (sum) {Summary};
            \draw[arr] (v) -- node[lbl, above] {\scriptsize\textsc{Done}} (sum);

            \draw[arr] (v.south) -- ++(0,-1.4) -| node[lbl, below, pos=0.3] {\scriptsize\textsc{Continue}} (p3.south);

            \path (input) -- (lit) coordinate[midway] (s0mid);
            \node[font=\tiny\bfseries, above=0.55cm of s0mid] {Stage 0};
            \path (p1) -- (v) coordinate[midway] (s1mid);
            \node[font=\tiny\bfseries, above=1.1cm of s1mid] {Stage 1: Proof Loop};
            \node[font=\tiny\bfseries, above=0.25cm of sum] {Stage 2};

            \node[font=\tiny\itshape, below=0.8cm of p3] {SV = Structural Verifier, DV = Detailed Verifier};
        \end{tikzpicture}%
    }
    \caption{QED system architecture. Stage~0 surveys the literature. Stage~1 iterates a proof loop: multiple models generate proofs in parallel; the structural verifier (SV) and detailed verifier (DV) each independently produce $n \times m$ verification reports ($m$ per proof), where $n$ is the number of provers and $m$ is the number of verification agents for each proof. At present, both $n$ and $m$ are at most three, corresponding to Claude Code, Codex, and Gemini CLI. A selector reads all verification results and picks the best proof with its verification result; and a verdict agent makes the final accept/continue decision based on the verification result of the selected proof. Stage~2 summarizes the entire proof effort. The prover can also access the internet when additional literature review is required.}
    \label{fig:architecture}
\end{figure*}

QED is a multi-stage, multi-agent pipeline for mathematical proof generation. Its architecture is directly motivated by the failure modes identified in Section~\ref{sec:failure_modes}: each design decision addresses a specific observed weakness. Figure~\ref{fig:architecture} provides an overview. We describe QED's two proof modes (simple and decomposition), its verification pipeline, and its multi-model capabilities.

\subsection{Simple Mode}
\label{sec:simple_mode}

QED simple mode (as shown in Fig~\ref{fig:architecture}) implements a three-stage pipeline, where Stage~0 performs literature survey and Stage~2 summarizes the entire proof effort. Stage~1 is essentially an iterative search--verify loop, suitable for problems that can be solved through progressive refinement.

\paragraph{Per-round flow in stage 1}
In each round $r = 1, \ldots, R_{\max}$, the system does the following steps:
\begin{enumerate}[leftmargin=*,itemsep=2pt]
    \item \textbf{Proof search}: One or more proof agents write a complete proof, informed by the literature survey. For rounds $r > 1$, the prover receives the previous round's proof and its verification results, so it can learn from past failures and attempt different strategies.

    \item \textbf{Verification}: Each proof is independently checked by structural verification (Phases 1--5: problem-statement integrity, completeness, citations, subgoal tree structure, and any human-specified rules) followed by detailed verification (Phase 6: step-by-step logical and mathematical checking).

    \item \textbf{Selection}: When multiple proofs are generated, a selector agent evaluates all proof--verification pairs and selects the best candidate.

    \item \textbf{Verdict}: A verdict agent reads the selected proof and its verification reports and decides \textsc{Done} (proof accepted) or \textsc{Continue} (next round). Note that the verdict agent does not read the proof or generate any new verification reports.
\end{enumerate}

\paragraph{Addressing Failure Mode 1 (context contamination).}
The proof agent and verification agents are separate model invocations with no shared context. The verifier sees only the proof document and the original problem---not the prover's reasoning process or intermediate attempts.

\paragraph{Addressing Failure Mode 7 (single-model bottleneck).}
When multi-model mode is enabled, multiple coding agents (e.g., Claude Code, Codex, Gemini CLI) generate proofs in parallel. Each proof is then independently verified by one or more verification agents, which may themselves use different models. A selector agent evaluates all proof--verification pairs and selects the best candidate. An optional brainstorm step allows multiple models to propose proof strategies before the main search begins.

\subsection{Decomposition Mode}
\label{sec:decomposition_mode}

For harder problems, QED offers a decomposition mode that separates proof planning from proof execution, directly addressing Failure Mode 4 (unstable proof plans).

\paragraph{Agents.} Decomposition mode employs six specialized agents:
\begin{enumerate}[leftmargin=*,itemsep=2pt]
    \item \textbf{Decomposer}: Creates a structured proof plan---a directed acyclic graph (DAG) of intermediate mathematical claims with explicit dependencies. It has access to previous plans and regulator output that identifies the failure reasons for the previous plans, allowing it to avoid repeating them or to revise them.

    \item \textbf{Single prover}: Writes a complete proof following the decomposer's plan. When the regulator decides \textsc{Revise\_Proof}, the prover receives the previous proof and its verification feedback to guide revision.

    \item \textbf{Structural verifier}: Phases 1–5 are the same as in the simple mode, except for an added criterion that checks whether the proof produced by the prover follows the decomposition plan.

    \item \textbf{Detailed verifier}: Phase 6 (same as the verification step in simple mode).

    \item \textbf{Regulator}: Analyzes verification failures and decides the next action.

    \item \textbf{Verdict}: Final accept/reject decision, same as the verdict agent in the simple mode.
\end{enumerate}

\paragraph{Three-level retry hierarchy.}
The regulator's key innovation is distinguishing three types of failure, each triggering a different corrective action:

\begin{table}[ht]
    \centering
    \small
    \begin{tabular}{@{}lll@{}}
        \toprule
        \textbf{Decision} & \textbf{Diagnosis} & \textbf{Action} \\
        \midrule
        \textsc{Revise\_Proof} & Execution errors, & Re-prove with \\
        & plan is sound & same plan \\
        \textsc{Revise\_Plan} & Structural gaps & Modify plan, \\
        & in the plan & then re-prove \\
        \textsc{Rewrite} & Fundamental & New decomposition \\
        & approach is flawed & from scratch \\
        \bottomrule
    \end{tabular}
    \caption{Three-level retry hierarchy in decomposition mode. The regulator distinguishes execution errors from plan-level and strategy-level failures. The decomposer-regulator design addresses Failure Mode 4.}
    \label{tab:retry}
\end{table}

Each level has a configurable limit (e.g., 3 proof attempts per plan revision, 2 plan revisions per decomposition, 3 total decompositions), preventing infinite loops while allowing systematic exploration.

\paragraph{Proof plan structure.}
The decomposer outputs a YAML-formatted plan where each step is a precise quantitative mathematical statement (not a vague description like ``show the function grows slowly''). Each step specifies its dependencies, difficulty level, and whether it is a \emph{key step}---the most novel and difficult part of the proof. The plan also includes cited sources and a mandatory self-critique section.

\begin{figure}[t]
    \centering
    \begin{tikzpicture}[
            node distance=0.45cm and 0.5cm,
            agent/.style={draw, rounded corners, fill=blue!10, minimum width=2.0cm, minimum height=0.6cm, align=center, font=\scriptsize},
            verifier/.style={draw, rounded corners, fill=orange!15, minimum width=2.0cm, minimum height=0.6cm, align=center, font=\scriptsize},
            regnode/.style={draw, rounded corners, fill=red!10, minimum width=2.0cm, minimum height=0.6cm, align=center, font=\scriptsize},
            decision/.style={draw, diamond, fill=yellow!15, minimum width=0.8cm, aspect=2, align=center, font=\scriptsize, inner sep=1pt},
            arr/.style={-{Stealth[length=4pt]}, semithick},
            loop/.style={-{Stealth[length=4pt]}, semithick},
            lbl/.style={font=\tiny, midway},
        ]
        \node[agent] (decomp) {Decomposer};
        \node[agent, below=of decomp] (prover) {Single Prover};
        \node[verifier, below=of prover] (sv) {Structural Verifier\\(Phases 1--5)};
        \node[verifier, below=of sv] (dv) {Detailed Verifier\\(Phase 6)};
        \node[regnode, below=of dv] (reg) {Regulator};
        \node[decision, below=0.6cm of reg] (dec) {Decision};

        \draw[arr] (decomp) -- node[lbl, right] {plan (YAML)} (prover);
        \draw[arr] (prover) -- node[lbl, right] {proof.md} (sv);
        \draw[arr] (sv) -- (dv);
        \draw[arr] (dv) -- node[lbl, right] {feedback} (reg);
        \draw[arr] (reg) -- (dec);

        \draw[arr] (sv.west) -- ++(-0.5,0) |- (reg.west);

        \draw[loop] (dec.east) -- ++(0.9,0) node[lbl, right, align=left, fill=white] {\textsc{Revise}\\[-1pt]\textsc{Proof}} |- (prover.east);

        \draw[loop] (dec.west) -- ++(-1.1,0) node[lbl, left, align=right, fill=white] {\textsc{Revise}\\[-1pt]\textsc{Plan}} |- ([yshift=-0.1cm] decomp.west);

        \draw[loop] (dec.south) -- ++(0,-0.35) -| (-2.2,0.0) node[lbl, left, pos=0.6, fill=white] {\textsc{Rewrite}} |- ([yshift=0.1cm] decomp.west);

        \node[font=\scriptsize\bfseries, right=1.8cm of reg] (done) {DONE};
        \draw[arr] (dv.east) -- ++(0.3,0) |- (done);

    \end{tikzpicture}
    \caption{Decomposition mode flow. The regulator analyzes verification failures and selects one of three corrective actions: \textsc{Revise\_Proof} revises the proof without changing the plan, \textsc{Revise\_Plan} modifies the existing decomposition plan, and \textsc{Rewrite} works on a new decomposition entirely. The decomposer has access to previous plans to avoid repetition or make revisions.}
    \label{fig:decomposition}
\end{figure}
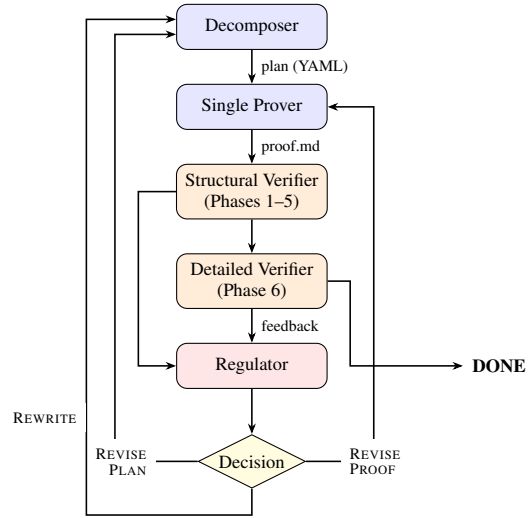

\subsection{Verification Pipeline}
\label{sec:verification}

QED's verification pipeline consists of six phases, split into two stages. This two-stage design addresses Failure Mode 5 (unfocused verification) by allowing each stage to focus on a specific class of checks.

\paragraph{Stage A: Structural Verification (Phases 1--5).}
Structural verification checks the proof's foundations before investing in expensive step-by-step analysis:

\begin{itemize}[leftmargin=*,itemsep=2pt]
    \item \textbf{Phase 1: Problem-statement integrity} (addresses Failure Mode 6). The verifier reads the original problem verbatim and compares it word-by-word against the claim the proof actually proves. \emph{Any} discrepancy---changed quantifiers, strengthened hypotheses, weakened conclusions, restricted domains, swapped conclusions and hypotheses---is an automatic failure.

    \item \textbf{Phase 2: Completeness and originality.} Checks that all parts of the problem are addressed and that the proof contains genuine argument (not just a literature survey or proof sketch).

    \item \textbf{Phase 3: Citation verification} (addresses Failure Mode 2). For every citation the verifier independently checks: (a) the source URL resolves, (b) title and authors match, (c) the exact statement exists at the specified location, (d) the cited statement matches the source verbatim, and (e) the proof correctly applies the cited result with all conditions satisfied.

    \item \textbf{Phase 4: Subgoal tree validation.} The proof must declare its logical architecture as a tree of subgoals rooted at the main claim. The verifier checks well-formedness (no orphans, no cycles), validity of each reduction, and completeness (every subgoal resolved, every cited theorem's conditions accounted for as explicit subgoals).

    \item \textbf{Phase 5: Human-specified rules.} If domain experts have provided additional verification criteria, these are checked as hard requirements.
\end{itemize}

\paragraph{Stage B: Detailed Verification (Phase 6).}
Phase 6 runs \emph{only if} Phases 1--5 pass. It performs step-by-step logical and mathematical checking:
\begin{itemize}[leftmargin=*,itemsep=2pt]
    \item For each assertion: state the claim, quote the justification, list dependencies, check logical validity, verify computations, and assign a verdict (Pass/Fail/Uncertain).

    \item Cross-reference with citation verdicts from Phase 3.

    \item Verify all subgoal resolutions point to specific, real parts of the proof.

    \item Analyze \texttt{<key-original-step>} tags (addresses Failure Mode 3): check that every tagged step is genuinely nontrivial and maximally detailed, and that every nontrivial step is properly tagged.
\end{itemize}

\subsection{Prover Output Format Enforcement}
\label{sec:key_step}

In both simple mode and decomposition mode, QED imposes a structured output format on the prover to address multiple failure modes simultaneously. To address Failure Mode 2 (citation hallucination), the prover must use a standardized citation format: every cited result must include the exact theorem statement, the source (with title, authors, and location), and a verification that all conditions of the cited result are satisfied in the current context. To address Failure Mode 3, QED requires the prover to wrap every nontrivial original step with \texttt{<key-original-step>} tags. Content inside these tags must be ``maximally detailed''---every sub-step, every bound, every case made explicit. The verifier then checks:
\begin{itemize}[leftmargin=*,itemsep=2pt]
    \item Every nontrivial proof has at least one key-original-step.

    \item Tagged content contains no vague language (``clearly,'' ``obviously,'' ``straightforward'').

    \item Untagged nontrivial steps are flagged as potential hidden weaknesses.

    \item Routine steps inflated with key-original tags are flagged.
\end{itemize}

This creates a forcing function: the prover must \emph{identify} the hardest parts of the proof and provide complete detail for exactly those parts.

\subsection{Optional Human-in-the-Loop Steering}
\label{sec:human_loop}

While QED runs fully automatically by default, it supports optional human steering between rounds. Domain experts can provide:
\begin{itemize}[leftmargin=*,itemsep=2pt]
    \item \textbf{Global proof hints}: Persistent guidance across all rounds (e.g., ``try Fourier-analytic methods'').

    \item \textbf{Additional verification rules}: Extra criteria the verifier should enforce.
\end{itemize}
These are read from designated files before each new round begins. In our evaluation on open research problems (Section~\ref{sec:open_problems}), this feature was \emph{not used}---all results were obtained without expert guidance.

\subsection{Standalone Proof Verifier}
\label{sec:standalone_verifier}

In addition to the full proof-generation pipeline, QED provides a \emph{standalone proof verifier} that takes any problem--proof pair and produces a structured verification report. This tool is designed for users who already have a proof (whether human-written or AI-generated) and want to check its correctness without running the full proof-search loop.

The standalone verifier uses a difficulty-adaptive architecture with up to three agents:
\begin{enumerate}[leftmargin=*,itemsep=2pt]
    \item \textbf{Difficulty judge.} A judge agent classifies the problem as \emph{Easy} or \emph{Hard}. For easy problems, the judge performs a complete streamlined verification in a single pass and the pipeline terminates. For hard problems, the judge defers to two specialized verification agents.

    \item \textbf{Structural verifier} (hard only). Checks problem-proof alignment, completeness and originality, and the proof's logical architecture.

    \item \textbf{Detailed verifier} (hard only, after structural pass). Performs step-by-step logical verification, mathematical correctness checks, rigor analysis, and coverage checks. Skipped if structural verification fails.
\end{enumerate}

\subsection{Automatic Resume}
\label{sec:resume}

QED supports automatic resume from any point in the pipeline. If the process exits for any reason---model API failures, network interruptions, machine restarts, or resource limits---QED detects the exact stage of progress by scanning the output directory structure and resumes from where it left off. This applies to both simple and decomposition modes: in simple mode, QED can resume at any sub-stage within any round; in decomposition mode, it restores the full attempt--revision--proof hierarchy and continues from the last incomplete step. This is essential for research-level proof tasks, which may run for extended periods across many rounds.

\section{Experiments}
\label{sec:open_problems}

We evaluate QED on 18 open research projects contributed by domain experts, with no proof known for any of the problems in those projects.
Our evaluation takes the form of case studies rather than benchmark experiments.
We do not report full details of all the unsuccessful cases for QED because some of the research problems remain active, with human mathematicians still working on them.
We also do not include full ablation studies over all projects due to two considerations.
First, no benchmark exists for open research problems.
By definition, these problems lack known solutions against which to measure performance.
Second, verifying the correctness and originality of a proof requires substantial effort from domain experts.
Running QED with each component disabled would multiply the number of proofs requiring expert review, imposing an impractical burden on the mathematicians who volunteered their time.
Instead, Section~\ref{sec:discussion} reports targeted diagnostic ablations on the two probability cases where repeated system variants could be compared without requiring broad expert re-review.
In practice, for each non-probability problem, we asked the corresponding expert to assess the correctness of QED's proof at most once; if the proof was found to be incorrect, we did not repeatedly revise and resubmit proofs for the same problem, in order to respect the experts' limited availability and avoid overburdening them.

\subsection{Experiment Setup}

\paragraph{Research Projects.}
18 research projects were contributed by 7 (groups of) experts. Each project may contain 1 or more open problems.
\begin{itemize}[leftmargin=*,itemsep=2pt]
    \item \textbf{Projects 1--10}: Contributed by a domain expert in probability. Some of the problems in the projects arose from domain expert's active research, and some of them are well-known, genuinely difficult open problems in the area.

    \item \textbf{Projects 11--12}: Contributed by a domain expert in operator algebra. Both of them are well-known, genuinely difficult open problems in the area.

    \item \textbf{Projects 13--14}: Contributed by a domain expert in algebraic geometry. All problems in the projects arose from domain expert's active research.

    \item \textbf{Projects 15--17}: Contributed by a domain expert in fluid PDE. All problems in the projects arose from domain expert's active research.

    \item \textbf{Project 18}: Contributed by a group of domain experts in inverse problems, concerning Carleman weight functions for PDEs. The statement and the proof are withheld currently.
\end{itemize}

The 5 research projects that QED successfully solved independently are documented in Appendices~\ref{app:pos1}--\ref{app:pos5}.
They correspond respectively to Project~6 (probability), Project~8 (probability), Project~13 (algebraic geometry), Project~15 (fluid PDE), and Project~18 (inverse problems).
Selected unsuccessful projects are presented in Appendix~\ref{app:neg}.

\paragraph{Configuration and Costs.}
The model configurations differed by projects:
\begin{itemize}[leftmargin=*,itemsep=2pt]
    \item \textbf{\ref{app:pos1}} Decomposition mode. All agents use Codex with GPT-5.5-pro, thinking level xhigh. This case costs around 1000 US dollars.

    \item \textbf{\ref{app:pos2}--\ref{app:pos3}} Decomposition mode. All agents use Codex with GPT-5.5, thinking level xhigh. \ref{app:pos2} costs around 200 US dollars. \ref{app:pos3} costs around 100 US dollars.

    \item \textbf{\ref{app:pos4}} There are three open research problems in this case. The first problem is completed with simple mode, with all agents using Codex with GPT-5.4, thinking level xhigh. The second and third problems are completed with decomposition mode, with all agents using Codex with GPT-5.5, thinking level xhigh. This case costs around 600 US dollars.

    \item \textbf{\ref{app:pos5}} Simple mode. Literature survey, proof search, all verifications and verdict use Anthropic Claude Opus 4.6, brainstorm is not enabled. This case costs around 50 US dollars.
\end{itemize}

We report the retry details, including the number of plan rewrites, plan revisions, and proof revisions, in Appendix~\ref{app:retry}.

\paragraph{Independence.}
The system received only the problem statements and configuration parameters; no domain-specific guidance or expert feedback was provided.

\subsection{Results}
\paragraph{Expert Assessment of Significance}
\label{res:expert_significance}

To assess the mathematical significance of the results produced by QED, we asked the collaborating domain experts to evaluate the completed solutions in the context of their respective fields. The experts confirmed that the resulting proofs are correct, original, and nontrivial. They further characterized the significance of the individual contributions as follows: \ref{app:pos1} was described as comparable to work published in the \emph{Electronic Journal of Probability} or the \emph{Proceedings of the American Mathematical Society}; \ref{app:pos2} was characterized as PhD-level work with nontrivial technical depth; the proof in \ref{app:pos3} was described as original, nontrivial, elementary, elegant, and ``moderately difficult''; \ref{app:pos4} was described as comparable to work published in venues such as the \emph{Journal of Differential Equations} or \emph{SIAM Journal on Mathematical Analysis}; and \ref{app:pos5} was characterized as comparable to work published in journals such as the \emph{Journal of Mathematical Analysis and Applications}. These assessments suggest that QED is not merely generating formal exercises or routine derivations, but is capable of producing solutions in diverse math domains, whose mathematical content reaches the level of publishable research contributions.

See appendix~\ref{app:pos} for the statements, proof links, and full expert comments.
\paragraph{Verifier Reliability}
\label{res:verifier-reliability}

In the QED pipeline, each proof candidate generated by the QED prover is checked by the QED verifier. We analyze the subset of QED runs in which the verifier used the Codex GPT-5.5 configuration. This subset contains 214 proof candidates spanning all 18 research projects considered in this work.

Among these 214 proof candidates verified by the QED verifier with Codex GPT-5.5, 17 were accepted by the verifier. Each of these 17 verifier-accepted proofs was subsequently reviewed and accepted by the corresponding domain expert. Thus, within the proof candidates checked under this verifier configuration, we observed no false positives: there was no case in which the verifier accepted a proof candidate that was later rejected by expert review.

\section{Discussion}
\label{sec:discussion}

\paragraph{Beyond benchmarks.}
These results demonstrate a capability qualitatively different from the benchmark performance discussed in \Cref{sec:related_work}.
Solving a competition problem with a known solution is fundamentally different from producing an original proof for a problem whose answer is not known in advance, which is why the evaluation in \Cref{sec:open_problems} is organized as expert-assessed research case studies rather than as benchmark scoring.

\paragraph{Originality.}
A central concern about AI-generated mathematics is whether the resulting proofs are genuinely original.
We argue that the proofs produced by QED for the five research projects above satisfy this criterion.
The problems addressed by QED were open research questions with no known solutions, as described in \Cref{sec:open_problems}; they were newly proposed by domain experts in the course of their own recent research.
QED produced all proofs without domain-expert guidance during the proof-generation process, consistent with the independence protocol in \Cref{sec:open_problems} and the unused human-steering option in \Cref{sec:human_loop}.
For Appendices~\ref{app:pos1}--\ref{app:pos5}, the corresponding experts characterized QED's contributions as original.

\paragraph{Targeted ablation studies.}
Full ablations are impractical for the reasons described in \Cref{sec:open_problems}: each candidate research proof requires time-intensive expert review.
We therefore performed targeted qualitative ablations on the two probability cases in Appendices~\ref{app:pos1} and~\ref{app:pos2}.
These two cases were selected because both were solved successfully by QED and expose the failure modes that motivated the system design in \Cref{sec:failure_modes,sec:system}.
As a single-query baseline, we submitted each problem to GPT-5.5 Pro in five independent attempts. None of the ten attempts produced a complete proof; all omitted essential steps.
We then inspected variants in which specific QED constraints were removed.
The goal of these ablations is not to estimate a success rate, but to test whether individual system components prevent the concrete failure modes identified in \Cref{sec:fm2,sec:fm3,sec:fm4}.

\begin{itemize}[leftmargin=*,itemsep=2pt]
    \item \textbf{Standardized citation format.}
    Without the format required in \Cref{sec:key_step}, the prover cited results without stating them precisely.
    In some attempts, it misstated a cited theorem, making the claimed application invalid. In others, the application was correct but insufficiently explained, falling short of mathematical writing standards and making the argument difficult for human experts to verify.
    This supports the claim from \Cref{sec:fm2,sec:key_step} that citation structure is not merely cosmetic: it forces cited results to be explicit enough for both verifier checking and expert review.

    \item \textbf{Citation verification (Phase 3).}
    Without independent citation verification, one attempt used a theorem found only in unpublished research notes.
    As a result, a key step in the proof relied on a less trustworthy source that readers could not independently verify.
    This supports the role of Phase~3 in \Cref{sec:verification} and directly addresses the citation hallucination failure mode in \Cref{sec:fm2}: the verifier must check not only whether a statement is mathematically plausible, but whether the cited source is public, precise, and usable in the proof.

    \item \textbf{\texttt{<key-original-step>} tags.}
    Both probability questions require matching lower and upper bounds.
    Without mandatory tagging, the prover developed the easier lower bound but left the harder upper bound as a sketch.
    This illustrates the misallocation of proof effort described in \Cref{sec:fm3}.
    It also supports the purpose of \texttt{<key-original-step>} tags in \Cref{sec:key_step} and their verification role in \Cref{sec:verification}: they force the prover and verifier to identify the mathematically novel step and demand maximal detail precisely there.

    \item \textbf{Decomposition mode.}
    Neither question was solved completely in simple mode, whose iterative search--verify loop is described in \Cref{sec:simple_mode}.
    Across repeated attempts, the prover frequently changed its overall proof strategy, indicating that the proof plan was unstable in the sense of \Cref{sec:fm4}.
    As a result, no single approach was developed in sufficient depth. The manuscripts produced in simple mode were at most 10 pages long.
    By contrast, decomposition mode produced the expert-accepted proofs in Appendices~\ref{app:pos1} and~\ref{app:pos2}, which were 20 and 15 pages, respectively. Although length alone does not measure proof quality, the qualitative difference was that decomposition mode preserved a stable plan long enough to resolve the difficult upper-bound arguments.
    This supports the claim in \Cref{sec:decomposition_mode,tab:retry} that separating planning from execution helps address unstable proof plans.
\end{itemize}

These ablations are limited to two cases and are diagnostic rather than statistically powered.
Nevertheless, they show that the tested components address distinct, directly observed failure modes from \Cref{sec:failure_modes}: citation grounding, verification of external dependencies, allocation of effort to key original steps, and stability of the proof plan.

\section{Conclusion}

We presented QED, an open-source multi-agent system that fully automates theorem proving given human-provided questions. Through collaborations with domain experts, QED produced original and nontrivial proofs for five research projects across probability, algebraic geometry, fluid PDEs, and inverse problems, with three of them comparable in difficulty and scope to work commonly published in established specialist mathematics venues. In each successful case, the system received only the problem statement and configuration parameters, ran without human guidance during proof generation, and produced outputs subsequently validated by the relevant experts. These results provide evidence that QED has a meaningful success rate for producing original and nontrivial proofs in mathematical research.

\begin{acks}
    We thank Xiaoqian Xu (Assistant Professor of Mathematics at Duke Kunshan University and Zu Chongzhi Center for Mathematics and Computational Sciences) for contributing Projects 15--17 and providing expert verification and commentary.
    We thank Yilong Zhang (Limited Term Assistant Professor in the Department of Mathematics at the University of Georgia) for contributing Projects 13--14 and providing expert verification and commentary.
    We thank Qiao~Zhuang (Assistant Professor in the Department of Mathematics and Statistics at University of Missouri--Kansas City), Zhongqiang Zhang (Associate Professor in the Department of Mathematical Sciences at Worcester Polytechnic Institute), and Yu Wang (Assistant Professor in the School of Mathematics at Southwest Jiaotong University) for contributing Project 18 and providing expert verification and commentary.
    We thank Kewei Xu (Kai Toyosawa, Postdoctoral Researcher in the Department of Mathematics, University of Münster) for contributing Projects 11--12 and providing expert verification.
    We thank Sam Buss (Professor at University of California, San Diego) for his continued support and insightful discussions.
\end{acks}

\bibliography{ref}
\bibliographystyle{icml2026}

\appendix

\section{Successful Projects}
\label{app:pos}

Table~\ref{tab:appendix-proof-index} collects the public proof records, expert comments, and withheld items for the five successful projects.

\begin{table*}[htp]
    \centering
    \footnotesize
    \caption{
        Appendix proof index for successful QED projects.
    }
    \renewcommand{\arraystretch}{1.1}
    \setlength{\tabcolsep}{5pt}
    \begin{tabularx}{\textwidth}{@{}p{0.12\textwidth}p{0.15\textwidth}X@{}}
        \toprule
        \textbf{Description} & \textbf{Domain} & \textbf{Record}\\
        \midrule
        Appendix~\ref{app:pos1}
        & Probability
        & Proof in~\cite{an2026returnprobabilityswitchwalkswitchlamplighter}; expert comments:
        {\scriptsize\url{https://github.com/proofQED/QED/blob/main/proved_statements/prob-May-15-2026/README.md}}\\

        Appendix~\ref{app:pos2}
        & Probability
        & Proof and expert comments in {\scriptsize\url{https://github.com/proofQED/QED/blob/main/proved_statements/prob-May-15-2026}}\\
        
        Appendix~\ref{app:pos3}
        & Algebraic geometry
        & Proof and expert comments in {\scriptsize\url{https://github.com/proofQED/QED/blob/main/proved_statements/algebraicgeometry-May-17-2026/problem-1-correct-proof.md}}\\

        Appendix~\ref{app:pos4}
        & Fluid PDE
        & Proof in~\cite{an2026lowerboundsadvectiondiffusionequations};
        expert comments:
        {\scriptsize\url{https://github.com/proofQED/QED/blob/main/proved_statements/analysis-May-19-2026/README.md}}\\

        Appendix~\ref{app:pos5}
        & Inverse problems
        & Withheld until related work is archived.\\
        \bottomrule
    \end{tabularx}
    \label{tab:appendix-proof-index}
\end{table*}

\subsection{Project~6 --- Return Probability Asymptotics for the Lamplighter Walk on $\mathbb{Z}_2 \wr T_d$}
\label{app:pos1}

\paragraph{Problem Statement.} Let $d \ge 3$ and let $T_d$ denote the infinite $d$-regular tree. Consider the switch--walk--switch lamplighter random walk on the wreath product $\mathbb{Z}_2 \wr T_d$, and let $p_{2n}(e,e)$ denote the return probability of this walk to the identity after $2n$ steps. Set
\begin{equation*}
    \rho_d \;=\; \frac{2\sqrt{d-1}}{d}.
\end{equation*}
Then the following sharp asymptotic holds:
\begin{equation*}
    p_{2n}(e,e)
    \;=\;
    \rho_d^{\,2n}\,
    \exp\!\left[
    -\bigl(\pi^{2}(\log(d-1))^{2} + o(1)\bigr)\,
    \frac{n}{\log^{2} n}
    \right].
\end{equation*}

\paragraph{Workflow.} The domain expert provided the problem statement to QED with no further mathematical input. QED was run in decomposition mode and produced the correct proof through multiple rounds of refinement, including substantial changes of proof plan by the decomposition agent. The resulting proof was then verified by the expert.

\paragraph{Expert Comment}
This result is a solid specialized contribution in probability theory. Its proof has real mathematical content as it cleverly combines probabilistic constructions with spectral analysis. Its significance appears comparable to work suitable for venues such as the \emph{Electronic Journal of Probability} or \emph{Proceedings of the American Mathematical Society}. The question is in the style of precise probabilistic calculations and estimates.

\subsection{Project~8 --- Total Variation Asymptotics for the Switch--Walk--Switch Walk on $\mathbb{Z}_2 \wr \mathbb{Z}$}
\label{app:pos2}

\paragraph{Problem Statement.} Consider the discrete-time switch--walk--switch random walk on the wreath product $\mathbb{Z}_2 \wr \mathbb{Z}$, and for a starting point $x \in \mathbb{Z}_2 \wr \mathbb{Z}$ let $P_t^{x}$ denote the law of this walk at time $t$. Fix
\begin{equation*}
    x \;=\; (\mathbf{0},\,0)
    \qquad\text{and}\qquad
    y \;=\; (\mathbf{0},\,2).
\end{equation*}
Then the total variation distance between the two laws satisfies the sharp asymptotic
\begin{equation*}
    \bigl\| P_t^{x} - P_t^{y} \bigr\|_{\mathrm{TV}}
    \;\asymp\;
    t^{-1/2},
\end{equation*}
where $\asymp$ denotes equality up to positive multiplicative constants independent of $t$.

\paragraph{Workflow.} The domain expert provided the problem statement to QED with no further mathematical input. QED was run in decomposition mode and produced the correct proof through multiple rounds of refinement, including substantial changes of proof plan by the decomposition agent. The resulting proof was then verified by the expert.

\paragraph{Expert Comment}
This is a technically nontrivial Ph.D.-level problem in probability theory. The question again is in the style of precise probabilistic calculations and estimates.

\subsection{Project~13 --- Local Invariant Cycle Theorem over the Integers in Degree One}
\label{app:pos3}

\paragraph{Problem Statement.} Let $X \to B$ be a one-parameter family of complex projective varieties over a small disk $B$, and assume that the central fiber is a simple normal crossing divisor. Let $X_t$ denote a smooth nearby fiber, and let $T$ denote the monodromy operator acting on $H^{1}(X_t,\mathbb{Z})$. Then the restriction map
\begin{equation*}
    H^{1}(X,\mathbb{Z})
    \;\longrightarrow\;
    H^{1}(X_t,\mathbb{Z})^{T}
\end{equation*}
to the monodromy-invariant submodule is surjective. The classical (rational) invariant cycle theorem of Clemens--Schmid (1970s) establishes the analogous statement after tensoring with $\mathbb{Q}$; the integral version was open. In February 2026, Arapura--Greer--Zhang showed that the integral statement fails for $H^{2}$. The result above establishes the integral statement for $H^{1}$.

\paragraph{Workflow.} At the time QED was given this problem, the human author had already discovered a proof but had not yet released it publicly. The domain expert provided the problem statement to QED with no further mathematical input. The resulting proof was independently verified by the expert and was found to be mathematically different from the human author's proof. The human author subsequently posted their own proof to arXiv, which is included in~\cite{arapura2026failureinvariantcycletheorem}.

\paragraph{Expert Comment}
The method in the AI's proof that interprets the coboundary map $H^{1}(X^{*}) \to H^{2}(X, X^{*})$ as the residue map is original. This actually holds in higher degrees by working with Griffiths' residue, but in degree one this interpretation is the classical residue and is particularly useful for the integral local invariant cycle theorem, since one has the key lemma which is essential in the proof. The proof is elementary and elegant: the first lemma is essentially residue theory in complex analysis, and the second is essentially reduced to Picard--Lefschetz theory for curves. It provides a different point of view on the integral version of the invariant cycle theorem in degree one. The expert classified problem difficulty as follows: easy means that the expected results can be achieved by standard methods; moderately difficult means that the expected results are plausible, but the methods are not known in advance; and difficult refers to solving open problems that have remained unresolved for years, or finding unexpected examples or counterexamples. Under this expert classification, this problem is considered moderately difficult.

\subsection{Project~15 --- Explicit Lower Bounds for Advection--Diffusion Equations}
\label{app:pos4}

\paragraph{Problem Statement.} Consider the advection--diffusion equation
\begin{equation*}
    \partial_t \rho \;+\; \mathbf{u}\cdot\nabla\rho \;=\; \nu\,\Delta\rho
\end{equation*}
on the two-dimensional torus $\mathbb{T}^{2}$ with mean-zero smooth initial data $\rho_{0}\not\equiv0$ and divergence-free velocity field $\mathbf{u}$. The paper establishes explicit, constructive lower bounds on suitable norms of the solution in three settings.

\smallskip
\noindent\emph{Result 1 (Inviscid shear; polynomial $\dot H^{-1}$ lower bound).} Let $\theta$ solve the transport equation
\begin{equation*}
    \partial_t \theta \;+\; U(y,t)\,\partial_x\theta \;=\; 0,
    \qquad \theta(0)=\theta_{0},
\end{equation*}
with $U \in L^{\infty}_{t}\,W^{1,1}_{y}(\mathbb{T})$ and $\theta_{0}\in C^{\infty}(\mathbb{T}^{2})$ mean-zero. Then there exists a constant $c_{*}>0$, depending only on $\theta_{0}$ and $U$, such that
\begin{equation*}
    \|\theta(t)\|_{\dot H^{-1}} \;\ge\; \frac{c_{*}}{1+t^{2}}
    \qquad \text{for all } t\ge 0 .
\end{equation*}

\smallskip
\noindent\emph{Result 2 (Diffusive shear; exponential $L^{2}$ lower bound and uniform mixing scale).} Let $\rho$ solve
\begin{equation*}
    \partial_t \rho \;+\; U(t,y)\,\partial_x\rho \;=\; \nu\,\Delta\rho,
\end{equation*}
with $U\in L^{\infty}_{t,y}(\mathbb{T})$ and $\nu>0$. Then there are explicit constants $c_{2}, c_{*}>0$, computable from $\rho_{0}$, $U$, and $\nu$, such that
\begin{equation*}
    \|\rho(t)\|_{L^{2}} \;\ge\; \|\rho_{0}\|_{L^{2}}\, e^{-c_{2}t}
    \qquad (t\ge0),
\end{equation*}
and, in the small-diffusivity regime $0<\nu\ll1$, the mixing scale is uniformly bounded below:
\begin{equation*}
    \frac{\|\rho(t)\|_{\dot H^{-1}}}{\|\rho(t)\|_{L^{2}}} \;\ge\; c_{*}
    \qquad \text{for all } t\ge0 .
\end{equation*}

\smallskip
\noindent\emph{Result 3 (Fast time-periodic flows; exponential $L^{2}$ lower bound).} Let $u\in L^{\infty}_{t}\,W^{1,\infty}_{x,y}$ be divergence-free and $L$-periodic in time, and let $\rho$ solve
\begin{equation*}
    \partial_t \rho \;+\; u(At,x,y)\cdot\nabla\rho \;=\; \nu\,\Delta\rho .
\end{equation*}
Then there exist an explicit threshold $A_{0}>0$, exponent $c_{A}>0$, and constant $C>0$, all expressed in terms of $\rho_{0}$, $u$, $L$, and $\nu$, such that for every $A>A_{0}$,
\begin{equation*}
    \|\rho(t)\|_{L^{2}} \;\ge\; C\,e^{-c_{A}t}
    \qquad (t\ge0).
\end{equation*}
In particular, for time-independent flows and time-periodic flows of sufficiently high frequency, superexponential $L^{2}$ decay is impossible.

\paragraph{Workflow.} The domain expert formulated the problem statements but did not intervene in the construction of the arguments; no PDE-specific guidance was given during the proof-generation process. QED produced the complete chain of reasoning, from the reduction of each problem to its step-by-step estimates, and verified the logical consistency of the derivations. The expert confirmed that the QED outputs are correct.

\subsection{Project~18 --- Carleman Estimate Weight Functions}
\label{app:pos5}

\paragraph{Problem statement.}
Construct a smooth space--time weight function $\psi(x,t)$ (with auxiliary parameters) for the wave operator on a half-infinite domain such that it satisfies strict pseudoconvexity-type inequalities, positivity conditions, and sufficient growth at infinity to establish a valid Carleman estimate.
\emph{Note: the full problem statement and proof are withheld until the corresponding mathematics paper is posted on arXiv.}

\paragraph{Result.}
Using Claude Opus 4.6 for both proof generation and verification, QED (with simple mode) produced a candidate weight function and a complete proof in \textbf{Round 1}. The proof was produced without any domain-expert guidance except for the correction to the statement of problem 5 prompted by the AI's solution to the originally stated problem.
Subsequent expert review determined that this solution corresponded to a degenerate case, revealing that the original constraints were too weak (a formulation issue rather than a QED failure).
After strengthening the constraints, the experts instructed QED to continue from its previous candidate.
QED then returned a refined weight function together with complete proofs satisfying all strengthened conditions.
No human intervention was required beyond refinement of the problem specification.

\paragraph{Expert assessment.}
One of the most promising applications of QED to the theory of Carleman estimates is the automated search for admissible Carleman weights. The success of a Carleman estimate hinges on finding a weight function finely adapted to the operator, the geometry, the boundary structure, and the relevant pseudoconvexity conditions. Traditionally, this has been a highly problem-specific, experience-driven process, involving substantial trial and error and lengthy inequality verification. A specialized QED-style system could transform this step into a modular, parallelizable, and rigorously verifiable search procedure: candidate weights are generated, tested against the required structural conditions, iteratively refined, and ultimately checked with full rigor. Such a tool would be valuable not only for accelerating well-established constructions, but also for exploring problems where the correct ansatz has yet to emerge.

Experts believe that QED's contribution to this work is comparable to that of research published in journals such as the Journal of Mathematical Analysis and Applications.

\section{Unsuccessful Cases}
\label{app:neg}

The following problems were attempted by QED but were not solved within the system's preset budget.
Table~\ref{tab:appendix-unsuccessful-index} gives a compact index, followed by the full problem statements retained for future evaluation.

\begin{table*}[htp]
    \centering
    \footnotesize
    \caption{
        Index of unsuccessful QED cases disclosed in the appendix.
    }
    \renewcommand{\arraystretch}{1.1}
    \setlength{\tabcolsep}{5pt}
    \begin{tabularx}{\textwidth}{@{}p{0.12\textwidth}p{0.15\textwidth}X@{}}
        \toprule
        \textbf{Description} & \textbf{Area} & \textbf{Problem summary}\\
        \midrule
        Appendix~\ref{app:neg:p2}
        & Probability
        & Show that the invariant random proper edge-coloring measure on the $d$-regular tree is a factor of i.i.d.\\

        Appendix~\ref{app:neg:p3}
        & Probability
        & Show that every finitely dependent $\mathrm{Aut}(T_d)$-invariant process on a finite state space is a factor of i.i.d.\\

        Appendix~\ref{app:neg:p7}
        & Probability
        & Determine whether the three-dimensional self-avoiding walk has expected displacement larger than diffusive order, $d_3( n ) \gg n^{ 1 / 2 }$.\\

        Appendix~\ref{app:neg:p10}
        & Probability
        & Prove Morris' conjecture $p_c( T_4 ) = 1 / 2$ for zero-temperature Glauber dynamics on the $4$-regular tree.\\

        Appendix~\ref{app:neg:p11}
        & Operator algebra
        & Determine whether coarse embeddability is a $W^*$-invariant for countable discrete groups.\\
        \bottomrule
    \end{tabularx}
    \label{tab:appendix-unsuccessful-index}
\end{table*}

\subsection{Project 2}
\label{app:neg:p2}

\paragraph{Problem statement.}

Let $d\geq 3$. Let $T_d$ be the $d$-regular tree.
There is a unique $\mathrm{Aut}(T_d)$-invariant probability measure $\mu$ on
the set of proper $d$-colorings of $E(T_d)$, the set of edges of $T_d$.

This measure is easy to construct. Let $o$ be a fixed root. Among the edges in $T_d$ adjacent to $o$, choose the colors of the edges uniformly at random such that they are distinct. Then, working outwards independently, where every time there is a choice of $d-1$ distinct colors for $d-1$ edges to guarantee that the coloring is proper, choose the colors uniformly at random across the edges.

Prove that $\mu$ is a factor of i.i.d. Relevant definitions and properties can be found in Russell Lyons's \emph{Factors of IID on Trees}.

\subsection{Project 3}
\label{app:neg:p3}

\paragraph{Problem statement.}

Let $T_d$ be the $d$-regular tree for $d \ge 3$, with vertex set $V = V(T_d)$. We consider processes that are invariant under the full automorphism group $\mathrm{Aut}(T_d)$. Let $S$ be a finite state space (such as $\{0, 1\}$). For $K \subseteq V$, write $\sigma(K)$ for the $\sigma$-field on $S^V$ generated by the coordinate projections $\pi_x$ for $x \in K$.

Prove that if $\mu$ is an $\mathrm{Aut}(T_d)$-invariant, finitely dependent probability measure on $S^V$ (for $d \ge 3$), then $\mu$ is a factor of i.i.d. The definitions of factor of i.i.d.\ and finitely dependent are as below, which can all be found in (arXiv:1401.4197, pp.~3--4).

\begin{definition}[FIID]
    Let $L$ be Lebesgue measure on $[0, 1]$. We equip the domain space $[0,1]^V$ with the product measure $L^V$. The elements $\omega \in [0, 1]^V$ of the domain space are sometimes called \emph{labels}.

    A measurable map $\phi \colon [0, 1]^V \to S^V$ is called an \textbf{$\mathrm{Aut}(T_d)$-factor} if it is $\mathrm{Aut}(T_d)$-equivariant: for all $\gamma \in \mathrm{Aut}(T_d)$,
    \begin{equation*}
        \phi(\gamma \omega) = \gamma \big( \phi(\omega) \big) \quad \text{for } L^V\text{-a.e.\ } \omega \in [0, 1]^V.
    \end{equation*}
    Since the domain space is equipped with a product measure, or i.i.d., such a $\phi$ is called a \textbf{factor of i.i.d.}, or \textbf{FIID} for short. The push-forward measure $\phi_* L^V$ on $S^V$ is also called an FIID.
\end{definition}

\begin{definition}[Finite Dependence]
    An $\mathrm{Aut}(T_d)$-invariant probability measure $\mu$ on $S^V$ is \textbf{$m$-dependent} if $\sigma(K_1), \dots, \sigma(K_p)$ are independent whenever the sets $K_1, \dots, K_p \subseteq V$ are pairwise separated by graph distance strictly greater than $m$. We say that $\mu$ is \textbf{finitely dependent} if it is $m$-dependent for some $m < \infty$.
\end{definition}

\subsection{Project 7}
\label{app:neg:p7}

\paragraph{Problem statement.}

For each $n$, let $(X_i)_{0\leq i\leq n}$ be the simple nearest-neighbor random walk on $\mathbb{Z}^k$ conditional on $X_0=0$ and no self-intersections up to time $n$ --- that is, a self-avoiding walk. Let $d_k(n):=\mathbb{E}(\|X_n\|)$, the expected Euclidean distance of $X_n$ from the origin. Is it true that $d_3(n)\gg n^{1/2}$?

\subsection{Project 10}
\label{app:neg:p10}

\paragraph{Problem statement.}

Let $T_4=(V,E)$ be the infinite $4$-regular tree. Let
$(\sigma_0(v))_{v\in V}$ be i.i.d.\ with
\begin{equation*}
    \mathbb P_p(\sigma_0(v)=+1)=p,
    \qquad
    \mathbb P_p(\sigma_0(v)=-1)=1-p.
\end{equation*}
Run continuous-time zero-temperature Glauber dynamics: each vertex has an
independent rate-one Poisson clock, and when the clock at $v$ rings,
$\sigma(v)$ is updated to agree with the majority of its four neighbors;
in case of a tie, the new spin is chosen uniformly from $\{-1,+1\}$. Let $\mathbb P_p$ denote the law of the dynamics.

Define
\begin{equation*}
    \begin{aligned}
        p_c(T_4)
        &:=
        \inf \Big \{
        p\in[0,1]:\\
        &\mathbb P_p\left(
        \lim_{t\to\infty}\sigma_t(v)=+1
        \text{ for every }v\in V
        \right)=1
        \Big \}.
    \end{aligned}
\end{equation*}

Prove Morris' conjecture for $T_4$: for zero-temperature Glauber dynamics on the $4$-regular tree,
\begin{equation*}
    p_c(T_4)=\frac12.
\end{equation*}
Equivalently, for every $p>1/2$, the process fixates to $+1$ at every
vertex almost surely.

\subsection{Project 11}
\label{app:neg:p11}

\paragraph{Problem statement.}

Is coarsely embeddability a W$^*$-invariant? That is, if $\Gamma,\Lambda$ are countable discrete groups and the group von Neumann algebras $L\Gamma \simeq L\Lambda$ are isomorphic, then $\Gamma$ being coarsely embeddable into Hilbert spaces implies $\Lambda$ is coarsely embeddable into Hilbert spaces as well.

\section{Retry Information}
\label{app:retry}

In this section, \ref{app:pos1}--\ref{app:pos5} refer to the 5 successful QED projects.
The notation ``P'' refers to an internal problem statement within one of those appendices, not to the original research-project IDs from Section~\ref{sec:open_problems}.
For example, ``\ref{app:pos4}, P3'' is the third internal problem statement within \ref{app:pos4}, whereas Project~15 is the original research project corresponding to \ref{app:pos4}.

\begin{table}[h]
\centering
\caption{Per-attempt breakdown of proof generation runs in decomposition mode. \textbf{Attempt} = number of new plans generated. \textbf{Revisions} = same plan, number of times it is revised and the proof regenerated to completion. \textbf{Proof number} = same plan, number of times the proof is modified. Some successful appendix entries contain multiple internal problem statements, and not every internal problem statement necessarily appears in the final proof.}
\label{tab:qed-attempt-revision-proof}
\begin{tabular}{lccc}
\toprule
Run identifier & Attempt & Revisions & Proof number \\
\midrule
\ref{app:pos1} & 2 & 1 & 2 \\
\midrule
\ref{app:pos2} & 1 & 2 & 5 \\
\midrule
\ref{app:pos3} & 1 & 1 & 1 \\
\midrule
\ref{app:pos4}, P3 & 1 & 1 & 2 \\
\midrule
\ref{app:pos4}, P4 & 1 & 1 & 2 \\
\midrule
\ref{app:pos4}, P5 & 1 & 1 & 3 \\
\midrule
\ref{app:pos4}, P6 & 1 & 1 & 2 \\
\midrule
\ref{app:pos4}, P7 & 1 & 1 & 1 \\
\midrule
\ref{app:pos4}, P8 & 1 & 3 & 5 \\
\midrule
\ref{app:pos4}, P9 & 1 & 1 & 3 \\
\midrule
\ref{app:pos4}, P10 & 1 & 1 & 1 \\
\midrule
\ref{app:pos4}, P11 & 2 & 2 & 6 \\
\midrule
\ref{app:pos4}, P12 & 2 & 3 & 3 \\
\bottomrule
\end{tabular}
\end{table}

\begin{center}
\begin{minipage}{0.45\textwidth}
\captionof{table}{Simple/legacy-mode QED runs without decomposition, counted by verification rounds.}
\label{tab:qed-simple-rounds}
\centering
\begin{tabular}{lc}
\toprule
Run identifier & Verification rounds \\
\midrule
\ref{app:pos4}, P1 & 4 \\
\ref{app:pos4}, P2 & 4 \\
\ref{app:pos5} & 1 \\
\bottomrule
\end{tabular}
\end{minipage}
\end{center}

\end{document}